\documentclass[runningheads]{llncs}

\usepackage{eccv}

\usepackage{eccvabbrv}

\usepackage{graphicx}
\usepackage{booktabs}

\usepackage[accsupp]{axessibility}  

\usepackage{hyperref}

\usepackage{orcidlink}

\usepackage{xcolor}
\usepackage{enumitem}
\usepackage{wrapfig}
\begin{document}

\title{When Is a Task Vector Enough? An Empirical Theory of Implicit Multimodal ICL} 

\titlerunning{Empirical Theory of Implicit Multimodal ICL}

\author{Jiaqian Li\inst{1}\orcidlink{0009-0009-3391-8620}}

\authorrunning{J. Li}

\institute{Brown University}

\maketitle

\begin{abstract}
Implicit multimodal in-context learning compresses demonstrations into internal interventions, ranging from static task vectors to query-conditioned transformations and attention routing. Despite their common goal, these methods differ substantially in how the intervention depends on the query and where it modifies the model, leaving unclear which additional complexity is necessary for a given task. We propose the \emph{Selection--Realization Hypothesis}. It views demonstrations as inducing a compact family of internal changes from which the query selects, while the model's computation constrains how the selected change can be implemented. We evaluate this account using controlled multimodal tasks in which query dependence varies without changing the underlying task primitives or prompt format. By contrasting correct demonstrations with matched counterfactuals, we measure the structure of explicit M-ICL and test whether it predicts intervention behavior. We find that the success of a static task vector is closely tied to how much of the demonstration-induced change is shared across queries. Additional intervention complexity becomes useful when explicit M-ICL contains query-specific or distributed structure that a local additive shift cannot recover. These relationships extend to natural VQA benchmarks and support cost-aware method selection without access to test performance. Our results provide a unified empirical theory of when demonstrations can be compressed into a task vector and when a more expressive intervention is warranted.

\end{abstract}

\section{Introduction}
\label{sec}

Multimodal large language models can adapt to new tasks from a small set of
image--text demonstrations without updating their parameters. This capability,
known as multimodal in-context learning (M-ICL), enables a single model to
infer novel label mappings, follow task-specific output formats, and recognize
visual concepts introduced only through context
\cite{tai2023linkcontextlearningmultimodalllms,
doveh2024multimodalincontextlearningvision,
sun2024generativemultimodalmodelsincontext}. Its flexibility, however, comes
with substantial inference cost: multimodal demonstrations consume many
tokens and must be repeatedly encoded and attended to for every new query.
Implicit M-ICL addresses this limitation by compressing demonstrations into
internal interventions that can be reused without retaining the original
examples in the inference context
\cite{li2025implicitincontextlearning,multimodal_task_vectors,live,m2iv,mimic,icr}.

\begin{wrapfigure}{r}{0.57\columnwidth}
\vspace{-8pt}
\centering
\includegraphics[width=\linewidth]{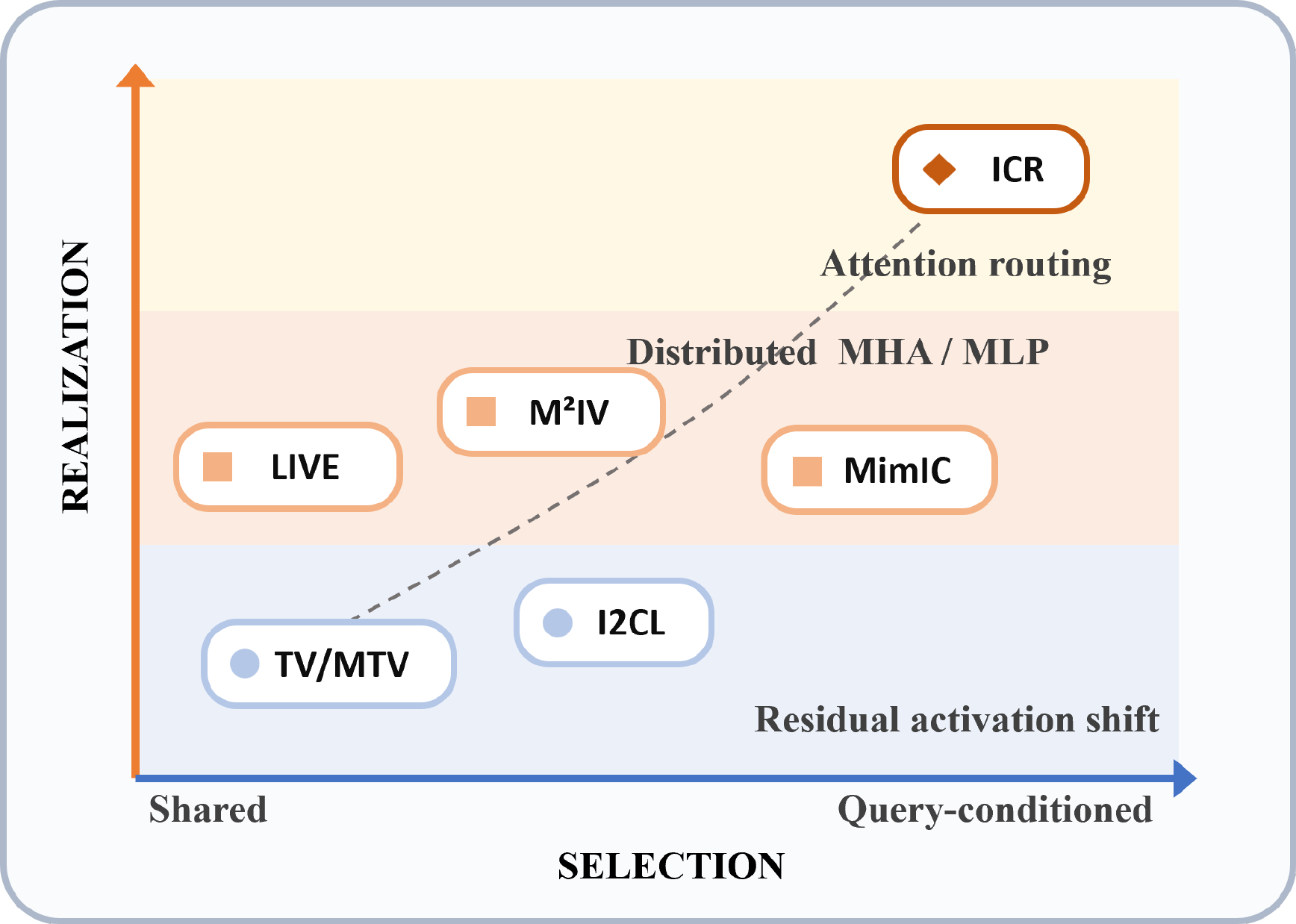}
\vspace{-6pt}
\caption{Conceptual design space of implicit M-ICL methods. Existing
approaches differ in whether intervention selection is shared or
query-conditioned and whether its realization uses local activation shifts,
distributed MHA/MLP modifications, or attention routing. The dashed curve
indicates increasing intervention expressivity rather than measured
performance.}
\label{fig:implicit_micl_space}
\vspace{-8pt}
\end{wrapfigure}

Existing implicit M-ICL methods realize this idea in markedly different ways.
Multimodal task vectors extract compact representations from selected
attention heads and reuse them across queries
\cite{multimodal_task_vectors}. LIVE and M$^2$IV distribute learned vectors
across attention or MLP components, allowing demonstration information to be
represented at multiple computational sites \cite{live,m2iv}. MimIC further
makes the magnitude of head-level shifts depend on the current query, whereas
ICR directly modifies attention logits through an input-conditioned router
\cite{mimic,icr}. These approaches therefore differ along two conceptually
distinct dimensions: how the demonstration-derived intervention is selected
for each query, and how that intervention is realized within the model.

Prior work shows that ICL can form compact task- or function-level representations that causally influence model behavior\cite{hendel2023taskvectors,todd2024functionvectors,liu2024incontextvectorsmakingcontext,multimodal_task_vectors}. Yet multimodal ICL remains sensitive to the composition and presentation of demonstrations, particularly the correspondence between visual and textual information\cite{qin2024factors,baldassini2024multimodal,stare}. These findings establish both the promise and the difficulty of compressing demonstrations, but they do not reveal which intervention design is actually necessary. Existing methods are usually evaluated as complete systems, so differences in intervention capacity, optimization, and supervision are entangled with the mechanism being tested. A query-conditioned method may outperform a static vector because the task genuinely requires query-dependent computation, but it may also benefit from a more flexible parameterization. The same ambiguity applies when comparing a local shift with a distributed or routing-based intervention. Downstream accuracy alone therefore cannot explain why additional complexity helps.

This paper asks whether the simplest adequate intervention can instead be predicted from the computation performed by explicit M-ICL. We call an intervention \emph{sufficient} when it recovers a fixed target fraction of the behavioral gain produced by explicit demonstrations. Among sufficient alternatives, the intervention with the lowest deployment cost is treated as \emph{minimal}. This definition shifts the problem from finding a universally strongest method to identifying how much complexity a particular demonstration-induced computation requires.

We propose the \emph{Selection--Realization Hypothesis} to provide such an
account. We hypothesize that a demonstration set defines a compact family of
internal transformations. The current image--text query determines which
transformation, or combination of transformations, is selected, while the
structure of the explicit in-context computation determines how it must be
realized inside the model. This view separates two sources of intervention
complexity. \emph{Selection complexity} describes whether the same
transformation is reused across queries or whether its composition must vary
with the input. \emph{Realization complexity} describes whether the selected
transformation can be implemented as a localized additive shift, requires
coordinated changes across attention and MLP components, or must alter
attention-mediated information flow. Static task vectors, multi-site vectors,
query-conditioned shifts, and attention routing then become nested cases of a
common intervention hierarchy rather than unrelated architectural choices. Figure~\ref{fig:implicit_micl_space} organizes representative implicit M-ICL
methods along these selection and realization dimensions, providing a unified
view of their otherwise heterogeneous designs.

We test the hypothesis by comparing the computations induced by correct demonstrations with those induced by matched counterfactuals. Controlled multimodal tasks allow query dependence to vary while the underlying task structure remains fixed. We find that demonstration-induced changes are compact but not always shared across queries. Their variation across queries and computational sites predicts both when a static vector fails and what additional flexibility is useful. The same diagnostics remain informative on natural tasks, where they support cost-aware method selection beyond the controlled setting.

Our study shifts the goal from finding the most expressive implicit M-ICL architecture to identifying the least costly intervention that preserves the behavior of explicit M-ICL. It offers an empirical theory in
which intervention complexity is predicted from measurable properties of the
demonstration-induced computation, rather than justified post hoc by benchmark
performance.

\section{A Unified Formulation of Implicit M-ICL}
\label{sec:formulation}

Let \(D=\{(x_i,y_i)\}_{i=1}^{m}\) denote a set of multimodal
demonstrations, and let \(x\) denote a new image--text query. A frozen model
with parameters \(\theta\) performs explicit M-ICL according to
\(p_\theta(y\mid x,D)\), whereas zero-shot inference uses
\(p_\theta(y\mid x)\). Implicit M-ICL instead compresses the demonstrations
into a cached representation \(S(D)\) and intervenes on the model while
processing the query:
\begin{equation}
p_{\mathcal{I}}(y\mid x,D)
=
p_\theta\bigl(y\mid x;\mathcal{I}(x,S(D))\bigr).
\label{eq:implicit_icl}
\end{equation}
The demonstrations are therefore absent from the inference context. The
objective is to recover their behavioral effect, without requiring every
intermediate computation to match explicit M-ICL.

To characterize that effect, let \(z_s(x;D)\) denote the computation recorded
at site \(s\). Depending on the intervention being studied, \(s\) may refer
to a residual state, a module output, or an attention-logit representation.
For every demonstration set, we construct matched counterfactuals
\(D_1^{-},\ldots,D_K^{-}\) that preserve the prompt structure while
disrupting the demonstrated input--output correspondence. We define the
resulting computation change as
\begin{equation}
\Delta z_s(x;D)
=
z_s(x;D)
-
\frac{1}{K}\sum_{j=1}^{K} z_s(x;D_j^{-}).
\label{eq:matched_effect}
\end{equation}
Using a matched context prevents changes caused only by prompt length or query
position from being attributed to the demonstrated mapping. The exact
construction of the counterfactuals is described in
Section~\ref{sec:framework}.

For a fixed \(D\), the collection of changes observed across queries defines
a demonstration-induced transformation,
\begin{equation}
\mathcal{T}_D(x)
\approx
\sum_{k=1}^{r}
c_k(x,D)\,\mathcal{B}_k(D),
\label{eq:operator_family}
\end{equation}
where \(\mathcal{B}_k(D)\) is a reusable basis element and \(c_k(x,D)\)
specifies how strongly it is expressed for the current query. Selection is
shared when these coefficients remain approximately constant across queries
and conditional when they vary systematically with \(x\). Realization is
determined by the computational objects represented by the basis elements and
the sites at which they act.

For an activation-based intervention at site \(s\), this formulation reduces
to
\begin{equation}
\Delta z_s(x;D)
\approx
v_s(D) + U_s(D)c_s(x,D),
\label{eq:additive_subspace}
\end{equation}
where
\begin{equation}
v_s(D)
=
\frac{1}{N}\sum_{i=1}^{N}\Delta z_s(x_i;D)
\label{eq:shared_component}
\end{equation}
is the component shared across the estimation queries. The columns of
\(U_s(D)\) span the remaining cross-query variation, and \(c_s(x,D)\) gives
the coordinates of the current query within this subspace. A static task
vector retains only the shared component, whereas a conditional intervention
also estimates the query-specific coordinates.

This formulation keeps compactness and static-vector sufficiency distinct. A
low-rank basis may reconstruct the changes on held-out queries even when those
queries occupy different points within the learned subspace. It also leaves
open how the recovered change must be realized inside the model.

\section{The Selection--Realization Hypothesis}
\label{sec:hypothesis}

The Selection--Realization Hypothesis begins with a necessary condition: the changes induced by a fixed demonstration set must exhibit reusable low-dimensional structure across queries. This structure allows us to ask how a query selects an appropriate transformation and how that transformation is instantiated within the model. We develop this account through four testable hypotheses.

\paragraph{H1: Demonstration-induced changes are compact.}
For a fixed demonstration set, the changes observed across queries should
occupy a space whose effective dimension is small relative to the maximum
dimension supported by the model width and the number of estimation queries.
A basis estimated from one group of queries should therefore reconstruct
changes on held-out queries. More importantly, intervening within this learned
space should recover more of the behavioral effect of explicit M-ICL than an
equally sized random subspace. 

\paragraph{H2: Sharedness and predictability determine intervention selection.}
Let $v_s(D)$ denote the shared component defined in
Section~\ref{sec:formulation}. We measure the fraction of
demonstration-induced energy that is shared across queries as

\begin{equation}
\mathrm{Shared}_s(D)
=
\frac{
    \left\|v_s(D)\right\|_F^2
}{
    \left\|v_s(D)\right\|_F^2
    +
    \frac{1}{N}
    \sum_{i=1}^{N}
    \left\|
        \Delta z_s(x_i;D) - v_s(D)
    \right\|_F^2
}.
\label{eq:sharedness}
\end{equation}

A static intervention should recover more of the effect of explicit M-ICL
when this quantity is high. However, low sharedness alone does not guarantee
that conditioning will help. The remaining coefficients must also be
predictable from the query without access to the demonstrations. We measure
this property using the held-out $R^2$ of a predictor that maps the
zero-shot query representation to $c_s(x,D)$. If recovery
depends on genuine query-specific selection, assigning otherwise valid
coefficients to the wrong queries should reduce performance.

\paragraph{H3: The distribution of causal support determines intervention location.}
A transformation may be shared across queries without being localized to a
single computational site. We estimate its causal support by measuring the
recovery obtained when matched additive interventions are applied separately
at candidate sites. When this support is concentrated, the strongest local
intervention should be sufficient. As support becomes more dispersed, a
multi-site intervention should provide a larger gain. This comparison holds
the selection mechanism and total intervention budget fixed, so any advantage
can be attributed to where the effect is realized rather than to additional
rank or parameters.

\paragraph{H4: An additive fit--recovery gap indicates an operator-level limitation.}
An additive intervention may reconstruct the representations produced by
explicit M-ICL without reproducing its behavior. We refer to this discrepancy
as the \emph{additive fit--recovery gap}. When additive reconstruction and
functional recovery are both high, modifying attention should offer little
additional benefit. When reconstruction remains high but recovery does not,
an attention-logit intervention may recover behavioral effects that activation
shifts miss. A routing gain that follows this gap under matched prediction and
parameter budgets would provide evidence that the limitation lies in how the
transformation is implemented, rather than in its dimensionality alone.

\section{Empirical Framework}
\label{sec:framework}

\begin{figure*}[t!]
\centering
\includegraphics[width=0.96\textwidth]{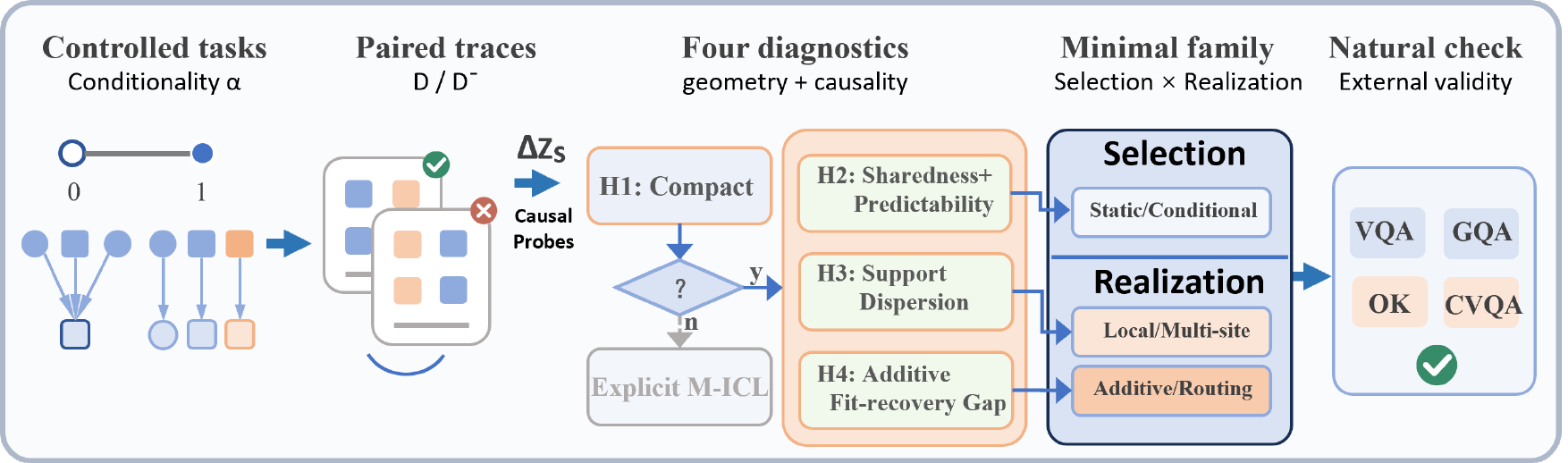}
\vspace{-2mm}
\caption{Overview of the empirical framework. Controlled episodes yield
paired computation changes $\Delta z_s$ from correct and counterfactual
demonstrations. H1 tests whether these changes admit compact compression; H2
determines whether selection can remain static or must depend on the query;
and H3--H4 determine whether the selected transformation can be realized
locally, across multiple sites, or through routing. Decision rules
calibrated on the controlled tasks are frozen before natural-task evaluation.}
\label{fig:empirical_framework}
\vspace{-3mm}
\end{figure*}

We first generate controlled episodes in which query dependence is varied while
the available task primitives and prompt format remain fixed. Paired traces
from correct and counterfactual demonstrations are then used to estimate the
four diagnostics in Section~\ref{sec:hypothesis}. Matched interventions test
whether these diagnostics identify the required selection and realization
mechanisms. Finally, a decision rule calibrated only on the controlled data is
frozen and evaluated on natural multimodal tasks. Figure~\ref{fig:empirical_framework} summarizes the empirical procedure. 

\subsection{Controlled Multimodal Task Family}

We construct episodes from synthetic scenes containing objects with
independently varied attributes and relations. Each episode samples a fixed
library of rules $\mathcal{R}=\{r_0,r_1,\ldots,r_J\}$ from the same set of
visual and textual primitives. Every query contains a gating attribute
$g(x)\in\{0,1\}$ and, when $g(x)=1$, a selector $q(x)\in\{1,\ldots,J\}$.
Both are recoverable from designated visual attributes or textual cues that
are present in every condition. The semantic answer is
\begin{equation}
a(x)=
\begin{cases}
r_0(x), & g(x)=0,\\
r_{q(x)}(x), & g(x)=1.
\end{cases}
\label{eq:controlled_rule}
\end{equation}
The first branch therefore reuses one rule across queries, whereas the second
requires the current query to select a rule. An episode-specific permutation
$\pi_D$ maps the semantic answer to an arbitrary single-token output symbol,
$y=\pi_D(a(x))$, so that the rule and output mapping must be inferred from
the demonstrations.

The conditionality level $\alpha\in[0,1]$ denotes the proportion of queries
assigned to the second branch. We use the same proportion in the
demonstrations and query splits of each episode, ensuring that the
manipulation changes selection complexity rather than introducing a
distribution shift. At $\alpha=0$, one rule is shared across all queries, and at
$\alpha=1$, every query selects its rule from the episode library.
Intermediate values produce controlled mixtures of these two cases. 

For every correct demonstration set $D$, we construct five counterfactual
sets $D_j^{-}$ by applying an independently sampled derangement to all
demonstration output symbols. The images, questions, example order, and token
lengths remain unchanged. Unlike an independent per-example shuffle, this
operation preserves a coherent episode-level mapping and the marginal label
distribution while making that mapping incorrect for the target episode.
We use these sets in Equation~\ref{eq:matched_effect} to remove effects caused
only by the presence and format of the demonstrations.

\subsection{Paired Traces and Diagnostic Estimation}

For each demonstration set and query, we record paired forward passes under
$D$ and $D_j^{-}$ at the final query token. The recorded sites include the
residual stream, MHA and MLP outputs, and attention logits across decoder
layers. Queries are partitioned before analysis into basis-estimation,
predictor-training, validation, and test splits. The test split is not used
to select ranks, sites, thresholds, or intervention hyperparameters.

H1 is tested by fitting the shared component and low-rank basis on the
estimation traces, with rank selected by the 90\% variance criterion. We
report effective rank and evaluate reconstruction and functional recovery on
held-out queries. A rank- and norm-matched random basis controls for low
dimensionality alone.

For H2, sharedness and held-out coefficient-prediction $R^2$ measure whether
query-varying structure can be selected from the zero-shot query. Static
and predicted conditional interventions use the same basis, while coefficient
shuffling tests whether recovery depends on the correct
query--transformation correspondence.

H3 and H4 concern realization rather than selection. To estimate causal
support, we apply the same matched additive probe separately at candidate
MHA and MLP outputs and form a site profile from their positive recovery
contributions. Its normalized entropy defines the support-dispersion score
$\kappa$. Residual-stream sites remain candidates for the localized
intervention but are not combined with their constituent MHA and MLP outputs
when computing $\kappa$. For H4, we measure the strongest additive
intervention's representational fit and behavioral recovery on held-out
diagnostic queries. Their difference,
\begin{equation}
G_{\mathrm{add}}
=
\mathrm{Fit}_{\mathrm{add}}
-
\mathrm{Recovery}_{\mathrm{add}},
\label{eq:additive_gap}
\end{equation}
is positive when an additive intervention reconstructs the explicit-M-ICL
representation more successfully than it recovers behavior. Routing gain is
evaluated on a disjoint test split, and its association with
$G_{\mathrm{add}}$ is tested while controlling for additive recovery. This
prevents the result from being driven solely by the additive term shared by
the two quantities.

\subsection{Matched Selection and Realization Tests}

The intervention comparison is factorial rather than a single nested
hierarchy. Along the selection axis, the static condition uses the shared
component for every query. The oracle conditional condition inserts the
coefficients recovered from each query's explicit-M-ICL trace and therefore
serves only as an upper bound on additive conditional sufficiency. The
deployable conditional condition instead uses coefficients predicted from
the zero-shot query representation.

Along the realization axis, the selected transformation is inserted at one
activation site, distributed across multiple MHA or MLP sites, or implemented
by modifying attention logits. Crossing the first two realization choices
with static and conditional selection gives the four additive families
evaluated later; attention routing is evaluated with the same query-side
predictor used by the conditional additive alternative.
This design allows H2 to vary selection while holding realization fixed, H3
to vary intervention location while holding selection fixed, and H4 to vary
operator type relative to the strongest matched additive intervention.

All comparisons use the same episodes and queries. Local and multi-site
interventions have the same total rank and injected norm, while predicted
conditional and routing variants are matched in trainable parameter and
supervision budgets. The random-basis and coefficient-shuffling controls test
H1 and H2, respectively. For H3 and H4, the relevant evidence is not a main
effect of greater expressivity, but whether the gain of the more complex
realization follows the corresponding diagnostic under these matched
budgets.

\subsection{Theory-Guided Selection and Natural-Task Validation}

The four diagnostics define a family selector without access to natural-task
test performance. A compactness gate first determines whether implicit
compression is supported by held-out reconstruction and recovery on a
calibration split. For compact transformations, sharedness and coefficient
predictability determine static versus conditional selection. Support
dispersion determines whether an additive transformation remains local or is
distributed across sites, and the additive fit--recovery gap determines
whether routing is admitted as a candidate. These decisions produce a set
$\mathcal{C}(D)$ of intervention families consistent with the measured
computation.

Let $d(D)$ denote the diagnostics measured from explicit-M-ICL calibration
traces. The selector chooses
\begin{equation}
\widehat{\mathcal{I}}(D)
=
\operatorname*{arg\,min}_{\mathcal{I}\in\mathcal{C}(D)}
\mathrm{Cost}(\mathcal{I})
\quad\text{subject to}\quad
\widehat{\mathrm{Recovery}}(\mathcal{I}\mid d(D))\geq\rho,
\label{eq:theory_selector}
\end{equation}
where $\rho$ is a sufficiency threshold fixed on the controlled validation
split before natural-task evaluation. The diagnostic thresholds and the
mapping from $d(D)$ to predicted recovery are calibrated on the same
controlled validation episodes and then frozen. If compactness fails or no candidate is
predicted to reach $\rho$, the selector abstains from implicit compression
and retains explicit M-ICL.

\section{Experiments}
\label{sec:experiments}

\subsection{Experimental Setup}
We evaluate OpenFlamingo-v2-9B~\cite{awadalla2023openflamingoopensourceframeworktraining},
Idefics2-8B~\cite{laurencon2024mattersbuildingvisionlanguagemodels}, and
LLaVA-NeXT-7B~\cite{liu2024improvedbaselinesvisualinstruction} on the
controlled task family and on VQAv2~\cite{goyal2017makingvvqamatter},
GQA~\cite{hudson2019gqanewdatasetrealworld},
OK-VQA~\cite{marino2019okvqavisualquestionanswering}, and
CVQA~\cite{romero2024cvqaculturallydiversemultilingualvisual}. All experiments use 16 shots. For each
$\alpha\in\{0,0.25,0.5,0.75,
1\}$, we sample 20 controlled episodes per
backbone, with four disjoint 100-query splits for basis estimation,
coefficient-predictor training, validation, and testing. Validation queries
are used to select ranks and intervention sites, fix diagnostic thresholds,
and perform early stopping. All reported results use only the test queries.
All conditional families use the same two-layer GELU selector (width 256),
which maps frozen zero-context query representations to $r$ intervention
coefficients and is trained by MSE with validation-based early stopping.
The intervention rank is the smallest rank explaining 90\% of estimation-set
variance, capped at 32. Coefficients are predicted from the zero-shot query
representation by a two-layer MLP with hidden size 128 and GELU activation.
We train it with AdamW for at most 1,000 steps, using learning rate
$10^{-3}$, weight decay $10^{-4}$, and early stopping on a held-out portion of
the predictor-training split. Natural-task experiments follow the 16-shot
protocol of M$^2$IV~\cite{m2iv}. For each model--dataset pair, diagnostics are
averaged over 20 demonstration sets and 256 calibration queries disjoint from
the evaluation queries. Intervention families are matched in total rank and
injected norm. Correlations and 95\% confidence intervals use 10,000
model--demonstration-set cluster-bootstrap samples. Runtime is measured on the
same GPU under an identical workload and includes extraction, prediction, and
diagnostic overhead.

\subsection{Evaluation Metrics}

\paragraph{Behavioral recovery.}
Let $\mathrm{Acc}_{D}$, $\mathrm{Acc}_{0}$, and
$\mathrm{Acc}_{\mathcal I}$ denote explicit-M-ICL, zero-shot, and
intervention performance. We normalize recovery as
\begin{equation}
\mathrm{Recovery}(\mathcal I)
=
\frac{\mathrm{Acc}_{\mathcal I}-\mathrm{Acc}_{0}}
{\mathrm{Acc}_{D}-\mathrm{Acc}_{0}}.
\label{eq:functional_recovery}
\end{equation}
We report this quantity only when explicit M-ICL exceeds zero-shot
performance by at least five percentage points. Recovery is not clipped. The
sufficiency threshold used by the family selector is fixed at $\rho=0.65$
before natural-task evaluation.

\paragraph{Selection diagnostics.}
H1 uses held-out reconstruction $R^2$ and the effective rank of the centered
demonstration-induced changes. For singular values $\sigma_j$,
\begin{equation}
r_{\mathrm{eff}}
=
\frac{\left(\sum_j\sigma_j^2\right)^2}{\sum_j\sigma_j^4}.
\label{eq:effective_rank}
\end{equation}
H2 uses sharedness from Equation~\ref{eq:sharedness}, held-out coefficient-
predictor $R^2$, and conditional gain
$\Delta R_{\mathrm{cond}}=R_{\mathrm{cond}}-R_{\mathrm{static}}$.
The controlled parameter $\alpha$ generates query dependence but is not used
by the selector.

\paragraph{Realization diagnostics.}
For each non-overlapping MHA or MLP site $s\in\mathcal S$, let
$\gamma_s=[\mathrm{Recovery}(\mathcal I_s)]_+$ and
$p_s=\gamma_s/\sum_{s'\in\mathcal S}\gamma_{s'}$. Support dispersion is
\begin{equation}
\kappa
=
-\frac{1}{\log|\mathcal S|}
\sum_{s\in\mathcal S}p_s\log p_s.
\label{eq:support_dispersion}
\end{equation}
Residual sites remain localized intervention candidates but are excluded from
$\kappa$ because they aggregate MHA and MLP outputs. H4 uses the additive
fit--recovery gap $G_{\mathrm{add}}$ from
Equation~\ref{eq:additive_gap}; routing gain is
$\Delta R_{\mathrm{route}}=R_{\mathrm{route}}-R_{\mathrm{add}}$.
Diagnostics are estimated from calibration queries and gains from disjoint
evaluation queries. The H3 and H4 correlations additionally control for
strongest-local and strongest-additive recovery, respectively.

\subsection{Results}

\paragraph{H1: Demonstration-induced changes are compact.}
The learned spaces transfer to held-out queries, whereas matched random spaces
do not (Table~\ref{tab:compactness}). 
\begin{wraptable}[13]{r}{0.48\columnwidth}
\vspace{-0.5em}
\centering
\tiny
\caption{Compactness of demonstration-induced transformations. L/R:
learned/random reconstruction; O/R: oracle/random recovery.}
\label{tab:compactness}
\setlength{\tabcolsep}{1.3pt}
\renewcommand{\arraystretch}{0.90}
\resizebox{\linewidth}{!}{%
\begin{tabular}{lccc}
\toprule
\textbf{Setting} & $r_{\mathrm{eff}}$
& \textbf{L/R Recon.} & \textbf{O/R Rec.} \\
\midrule
$\alpha=0$   & 1.1  & 0.88/0.01 & 0.95/0.02 \\
$\alpha=0.5$ & 5.8  & 0.82/0.02 & 0.91/$-0.01$ \\
$\alpha=1$   & 12.4 & 0.75/0.02 & 0.86/0.04 \\
VQAv2        & 18.5 & 0.68/0.04 & 0.78/0.05 \\
GQA          & 25.2 & 0.61/0.03 & 0.71/0.02 \\
OK-VQA       & 21.4 & 0.55/0.05 & 0.63/0.06 \\
CVQA         & 28.7 & 0.58/0.04 & 0.65/0.03 \\
\bottomrule
\end{tabular}}
\vspace{-0.8em}
\end{wraptable}
Effective rank rises with $\alpha$ and across natural tasks, but remains far
below model width. Learned bases also retain a clear advantage over random
bases in both reconstruction and recovery. Together, these results support
H1, although compactness does not imply that one direction is shared by all
queries.

\paragraph{H2: Sharedness and predictability determine static-vector sufficiency.}
Table~\ref{tab:controlled_interventions} reports the controlled sweep. Static
recovery declines sharply as more queries require conditional selection,
whereas the oracle conditional intervention remains effective. The predicted
intervention recovers much of this advantage, but shuffling its coefficients
across queries removes it. Thus the gain cannot be explained by rank or
activation magnitude alone.

\begin{table*}[t]
\centering
\small
\caption{Controlled results averaged over models and demonstration sets. Each
cell reports accuracy (\%) / normalized recovery.}
\label{tab:controlled_interventions}
\resizebox{\textwidth}{!}{%
\begin{tabular}{lcccccc}
\toprule
\textbf{Intervention}
& $\alpha=0$ & $\alpha=0.25$ & $\alpha=0.5$
& $\alpha=0.75$ & $\alpha=1.0$ & \textbf{Average} \\
\midrule
zero-shot
& 26.7 / 0.00 & 23.9 / 0.00 & 26.4 / 0.00
& 24.3 / 0.00 & 25.0 / 0.00 & 25.3 / 0.00 \\
Explicit M-ICL
& 91.1 / 1.00 & 89.2 / 1.00 & 88.6 / 1.00
& 85.4 / 1.00 & 84.5 / 1.00 & 87.8 / 1.00 \\
\midrule
Static additive
& 85.9 / 0.92 & 78.8 / 0.84 & 66.8 / 0.65
& 59.1 / 0.57 & 44.6 / 0.33 & 67.1 / 0.66 \\
Oracle conditional
& 87.9 / 0.95 & 84.0 / 0.92 & 83.0 / 0.91
& 78.1 / 0.88 & 76.2 / 0.86 & 81.8 / 0.90 \\
Predicted conditional
& 84.7 / 0.90 & 78.1 / 0.83 & 74.3 / 0.77
& 66.5 / 0.69 & 63.1 / 0.64 & 73.3 / 0.77 \\
Shuffled coefficients
& 82.1 / 0.86 & 70.9 / 0.72 & 55.6 / 0.47
& 46.9 / 0.37 & 33.9 / 0.15 & 57.9 / 0.51 \\
Random subspace
& 28.0 / 0.02 & 23.2 / $-0.01$ & 25.8 / $-0.01$
& 23.1 / $-0.02$ & 27.4 / 0.04 & 25.5 / 0.00 \\
\bottomrule
\end{tabular}}
\end{table*}

\begin{figure*}[t]
\centering
\includegraphics[width=0.98\textwidth]{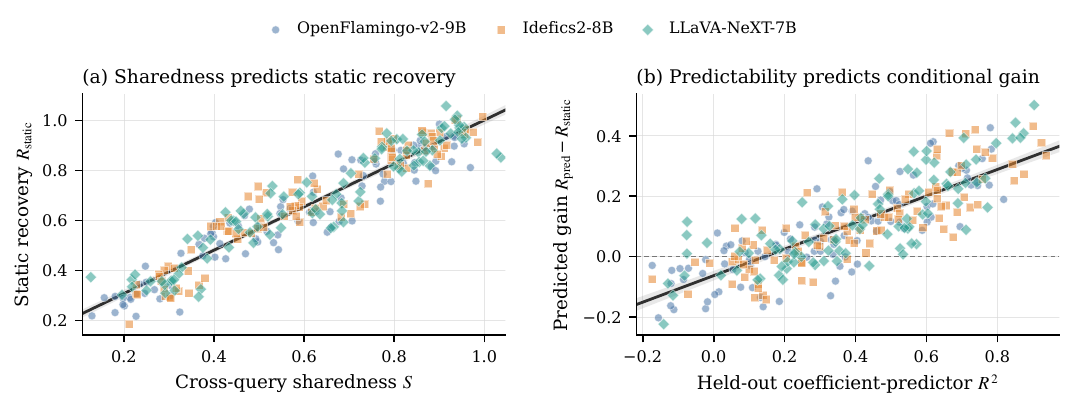}
\caption{Testing H2 on held-out queries. Each point is one
model--episode--condition observation. Panel (a) relates sharedness to static
recovery; panel (b) relates coefficient-predictor $R^2$ to the gain of
predicted conditioning. Lines show fitted trends with cluster-bootstrap
95\% confidence intervals.}
\label{fig:selection_results}
\end{figure*}

Figure~\ref{fig:selection_results} tests the measured diagnostics rather than
the generating parameter $\alpha$. Sharedness predicts static recovery
($\rho_s=0.96$), while held-out coefficient-predictor $R^2$ predicts the
gain from conditional selection ($\rho_s=0.80$). The remaining gap between
oracle and predicted conditioning shows why low sharedness alone is
insufficient: the query-varying coefficients must also be predictable.

\paragraph{H3--H4: Realization gains follow support and operator mismatch.}
Table~\ref{tab:realization_results} shows that increasing realization
complexity is not uniformly beneficial. With conditional selection held fixed,
distributing the same total rank and norm across multiple sites improves
recovery by $0.04$--$0.09$. Routing provides little additional benefit on the
controlled tasks and VQAv2, but exceeds the strongest additive intervention by
$0.04$--$0.06$ on GQA, OK-VQA, and CVQA.

\begin{table*}[t]
\centering
\small
\caption{Functional recovery of matched intervention families. Natural-task
results are averaged across the three LVLMs.}
\label{tab:realization_results}
\setlength{\tabcolsep}{5pt}
\renewcommand{\arraystretch}{1.05}
\resizebox{\textwidth}{!}{%
\begin{tabular}{lllccccc}
\toprule
\textbf{Intervention}
& \textbf{Selection}
& \textbf{Realization}
& \textbf{Controlled}
& \textbf{VQAv2}
& \textbf{GQA}
& \textbf{OK-VQA}
& \textbf{CVQA} \\
\midrule
Local static
& Static
& Single-site additive
& 0.66
& 0.64
& 0.49
& 0.43
& 0.45 \\

Local conditional
& Conditional
& Single-site additive
& 0.77
& 0.72
& 0.61
& 0.54
& 0.57 \\

Multi-site static
& Static
& Multi-site additive
& 0.71
& 0.65
& 0.56
& 0.50
& 0.52 \\

Multi-site conditional
& Conditional
& Multi-site additive
& 0.83
& \textbf{0.76}
& 0.69
& 0.63
& 0.66 \\

Attention routing
& Conditional
& Attention logits
& \textbf{0.84}
& 0.75
& \textbf{0.73}
& \textbf{0.68}
& \textbf{0.72} \\
\bottomrule
\end{tabular}%
}
\end{table*}

\begin{figure*}[t]
\centering
\includegraphics[width=0.98\textwidth]{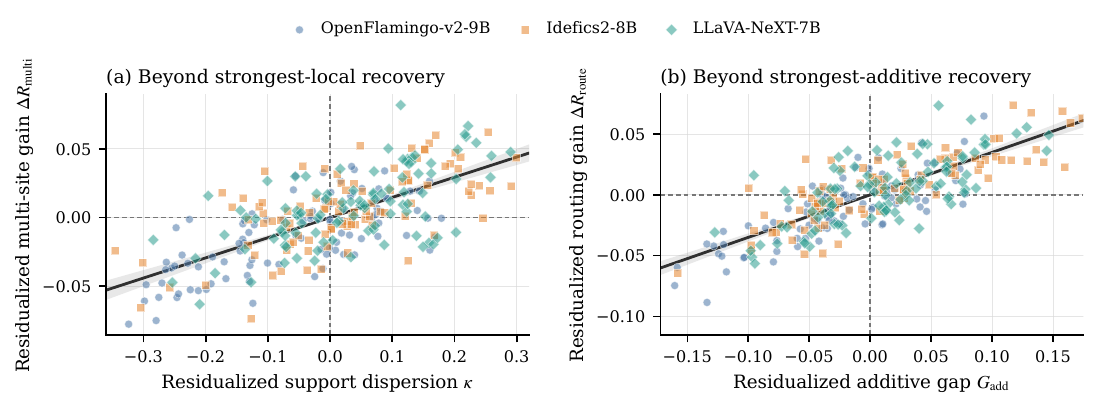}
\caption{Partial-residual tests of H3--H4. Panel (a) relates support
dispersion to multi-site gain after controlling for strongest-local recovery;
panel (b) relates the additive fit--recovery gap to routing gain after
controlling for strongest-additive recovery. Lines show fitted trends with
cluster-bootstrap 95\% confidence intervals.}
\label{fig:realization_diagnostics}
\end{figure*}

Figure~\ref{fig:realization_diagnostics} explains when these gains arise.
After controlling for strongest-local recovery, support dispersion remains
associated with multi-site gain
($\rho_{\mathrm{partial}}=0.67$), supporting H3. Likewise, the additive
fit--recovery gap predicts routing gain after controlling for
strongest-additive recovery
($\rho_{\mathrm{partial}}=0.77$). This supports H4 while showing that routing
is warranted specifically when additive realization is insufficient.

\paragraph{Natural-task validation and theory-guided selection.}
We freeze the diagnostic thresholds using only the controlled calibration
data. For each natural model--dataset pair, the selector determines an
admissible selection--realization family from calibration traces and deploys
the lowest-cost compatible method predicted to reach $\rho=0.65$.
Method-to-family assignments are fixed before evaluation, and the selector
does not observe natural-task test accuracy. Table~\ref{tab:natural_results} shows that the selector remains within
$0.21$--$0.37$ percentage points of the post-hoc best method, with mean
regret $0.29$. Its relative cost is $0.58$, a $19\%$ reduction from always
using M$^2$IV and a $42\%$ reduction from always using routing. Therefore, the theory
supports lower-cost method selection before test performance is
known, rather than merely identifying the most expressive intervention.

\begin{table}[t]
\centering
\small
\caption{Natural-task accuracy (\%) under the 16-shot protocol, averaged over
the three models. Relative cost is end-to-end GPU time normalized to ICR and
includes diagnostic overhead.}
\label{tab:natural_results}
\setlength{\tabcolsep}{3.8pt}
\resizebox{\columnwidth}{!}{%
\begin{tabular}{lcccccc}
\toprule
\textbf{Method}
& \textbf{VQAv2} & \textbf{GQA} & \textbf{OK-VQA}
& \textbf{CVQA} & \textbf{Avg.} & \textbf{Cost $\downarrow$} \\
\midrule
zero-shot
& 43.10 & 52.42 & 31.04 & 31.09 & 39.41 & -- \\
Explicit M-ICL
& 60.08 & 69.70 & 51.37 & 56.39 & 59.39 & -- \\
Task Vector~\cite{hendel2023taskvectors}
& 47.88 & 60.26 & 38.24 & 43.02 & 47.35 & 0.06 \\
I2CL~\cite{li2025implicitincontextlearning}
& 52.02 & 61.93 & 44.13 & 49.42 & 51.88 & 0.12 \\
LIVE~\cite{live}
& 62.22 & 69.20 & 53.47 & 55.42 & 60.08 & 0.70 \\
M$^2$IV~\cite{m2iv}
& \textbf{63.93} & \textbf{73.81} & 55.37 & 60.11 & 63.31 & 0.72 \\
MimIC~\cite{mimic}
& 63.13 & 72.94 & 54.87 & 60.76 & 62.93 & 0.84 \\
ICR~\cite{icr}
& 62.81 & 73.48 & \textbf{56.14} & \textbf{61.35} & 63.45 & 1.00 \\
\midrule
\textbf{Theory-selected}
& 63.56 & 73.51 & 55.93 & 61.09 & \textbf{63.52} & 0.58 \\
\bottomrule
\end{tabular}}
\end{table}

\section{Related Work}
\label{sec:related_work}

\paragraph{Implicit multimodal in-context learning.}
Representation-based approaches seek to reproduce the effect of
demonstrations without retaining them in the inference context. Early work
showed that task information can be extracted and transferred through
in-context, task, or function vectors
\cite{liu2024incontextvectorsmakingcontext,hendel2023taskvectors,
todd2024functionvectors}. I2CL compresses demonstrations into a context vector
and combines it with query activations during zero-shot inference
\cite{li2025implicitincontextlearning}. This paradigm has been extended to
multimodal models through attention-head task vectors, learned VQA
interventions, and vectors distributed across MHA and MLP components
\cite{multimodal_task_vectors,live,m2iv}. More expressive approaches introduce
input dependence through query-conditioned shift magnitudes, state-dependent
steering, or attention routing \cite{mimic,svf,icr}. 

\paragraph{Understanding in-context learning.}
ICL has been interpreted as implicit Bayesian inference, gradient descent, or
meta-optimization
\cite{xie2022explanationincontextlearningimplicit,
akyürek2023learningalgorithmincontextlearning,
vonoswald2023transformerslearnincontextgradient,
dai2023gptlearnincontextlanguage}. Mechanistic studies provide complementary
accounts in which induction heads retrieve context-associated patterns and
compact task or function representations causally influence model outputs
\cite{olsson2022incontextlearninginductionheads,
singh2024needsrightinductionhead,hendel2023taskvectors,
todd2024functionvectors}. Studies of multimodal ICL further show that behavior
depends on demonstration retrieval, ordering, modality balance, and prompt
construction \cite{qin2024factors,stare,baldassini2024multimodal,
chen2024multimodallargelanguagemodels}, while recent analyses identify
cross-modal circuits associated with label copying and task execution
\cite{huang2026dissectingmultimodalincontextlearning}.

\section{Future Directions}
\label{sec:future}

Our diagnostics could support a reusable library of interventions indexed by their selection and realization profiles, recovery and effective locations. For a new task, a small calibration set could retrieve or compose suitable entries, enabling more adaptive and efficient implicit M-ICL.

\section{Conclusion}
\label{sec:conclusion}

We present an empirical theory of implicit multimodal in-context learning that explains when a static task vector is sufficient and when more expressive interventions are required. Our account separates transformation selection from realization: static vectors suffice when demonstration-induced computation is shared across queries, query-conditioned interventions are needed when transformation coefficients vary predictably with the input, multi-site interventions address dispersed causal support, and routing is warranted when additive interventions fail to recover behavior. We evaluate these claims through controlled tasks, matched-capacity comparisons, causal controls, and natural multimodal benchmarks. By connecting measurable properties of explicit M-ICL to intervention success, our framework provides an evidence-based principle for selecting the minimal sufficient implicit M-ICL method.

\clearpage  

%
%
\bibliographystyle{splncs04}
\bibliography{main}
\appendix

\section*{Storyline}

\begingroup
\small
\setlength{\fboxsep}{1pt}
\newcommand{\firstterm}[1]{\colorbox{cyan!20}{#1}}
\newcommand{\reuseterm}[1]{\underline{#1}}
\setlist[enumerate]{nosep,itemsep=0pt,parsep=0pt,topsep=1pt}

\begin{enumerate}[leftmargin=1.4em,label=\arabic*.]
    \item \textbf{Why interesting?}
    \begin{enumerate}[leftmargin=1.6em,label=\alph*.]
        \item \firstterm{Explicit M-ICL}: demonstrations adapt a frozen model
        but are reprocessed for every query.
        \item \firstterm{Implicit M-ICL}: cache this effect; use the least
        expressive intervention that preserves \reuseterm{explicit M-ICL}.
    \end{enumerate}

    \item \textbf{How done now?}
    \begin{enumerate}[leftmargin=1.6em,label=\alph*.]
        \item \firstterm{Static task vectors} add one query-invariant shift;
        multi-site variants spread fixed shifts across components.
        \item Conditional methods use query-specific coefficients; routing
        changes computation paths.
        \item Comparisons often mismatch sites, capacity, training, or tuning.
    \end{enumerate}

    \item \textbf{What is missing, and So What?}
    \begin{enumerate}[leftmargin=1.6em,label=\alph*.]
        \item Gains confound query dependence, site/operator, and extra capacity.
        \item No diagnostic says when a \reuseterm{static task vector} suffices;
        the wrong choice loses accuracy or efficiency.
    \end{enumerate}

    \item \textbf{Proposed approach (P).}
    \begin{enumerate}[leftmargin=1.6em,label=\alph*.]
        \item Separate \firstterm{selection} (shared vs.\ query-specific) from
        \firstterm{realization} (where/how the transformation acts).
        \item \textbf{H1}: a compact transformation family exists; \textbf{H2}:
        shared selection implies static vectors, predictable variation implies
        conditional coefficients.
        \item \textbf{H3}: dispersed causal support requires multiple sites;
        \textbf{H4}: routing is needed only after additive shifts fit
        representations but fail behavior.
        \item Rank, coefficient predictability, support dispersion, and recovery
        gap select the \firstterm{minimal method family}.
    \end{enumerate}

    \item \textbf{Experimental questions.}
    \begin{enumerate}[leftmargin=1.6em,label=\alph*.]
        \item When is a task vector enough?
        \begin{enumerate}[leftmargin=2.7em,label=c\arabic*:]
            \item Shared, mixed, and fully query-conditioned episodes; vary the fraction of
queries requiring rule selection while holding task primitives and prompts fixed.
            \item Match examples, prompts, supervision, rank, and parameters;
            verify explicit M-ICL $>$ zero-shot.
            \item Static recovery increases with sharedness; true coefficients $>$
            shuffled coefficients.
            \item Conditional $>$ static/random iff coefficients are
            query-predictable.
        \end{enumerate}

        \item Where/how is intervention complexity needed?
        \begin{enumerate}[leftmargin=2.7em,label=c\arabic*:]
            \item Pair residual/attention/feed-forward/logit effects on the same
            episodes.
            \item Match rank, norm, layers, parameters, and coefficient predictor.
            \item Multi-site advantage follows support dispersion, not parameter
            count.
            \item Routing advantage requires an additive functional-recovery gap.
        \end{enumerate}

        \item Do the diagnostics generalize with natural confounders?
        \begin{enumerate}[leftmargin=2.7em,label=u\arabic*:]
            \item Models: Idefics2-8B, LLaVA-NeXT-7B, OpenFlamingo-v2-9B;
            datasets: VQAv2, GQA, OK-VQA, CVQA.
            \item Match 16-shot examples, queries, decoding, and supervision.
            \item Reproduce all four method families under one protocol.
            \item Test whether diagnostics predict recovery and the
            \reuseterm{minimal method family} across models and tasks.
        \end{enumerate}
    \end{enumerate}
\end{enumerate}

\endgroup

\end{document}